%% file: main.tex
\documentclass{sigkddExp}
\usepackage{url}
\usepackage{amsmath,amssymb}
\usepackage{xcolor}

\definecolor{myred}{rgb}{.8,.0,.0}

\DeclareMathOperator{\E}{\mathbb{E}}

\def\alignauthor{
\end{tabular}%
  
  \begin{tabular}[t]{p{\textwidth}}\centering}%
  
\begin{document}
%

\title{Predicting Bearings' Degradation Stages for\\
Predictive Maintenance in the Pharmaceutical Industry}
%

%


\author{
%

\alignauthor Dovile Juodelyte, Veronika Cheplygina, Therese Graversen, Philippe Bonnet\\
      \affaddr{IT University of Copenhagen}\\
      \affaddr{Copenhagen, Denmark}
      \email{\{doju, vech, theg, phbo\}@itu.dk}
}

\maketitle

\begin{abstract}
In the pharmaceutical industry, the maintenance of production machines must be audited by the regulator. In this context, the problem of predictive maintenance is not when to maintain a machine, but what parts to maintain at a given point in time. The focus shifts from the entire machine to its component parts and prediction becomes a classification problem. In this paper, we focus on rolling-elements bearings and we propose a framework for predicting their degradation stages automatically. Our main contribution is a k-means bearing lifetime segmentation method based on high-frequency bearing vibration signal embedded in a latent low-dimensional subspace using an AutoEncoder. Given high-frequency vibration data, our framework generates a labeled dataset that is used to train a supervised model for bearing degradation stage detection. Our experimental results, based on the FEMTO Bearing dataset, show that our framework is scalable and that it provides reliable and actionable predictions for a range of different bearings.
\end{abstract}

\section{Introduction} \label{sec:introduction}

\input{sec1_introduction}

\section{Related Work}

\input{sec5_related}

\section{Domain Knowledge} \label{sec:problem}
\input{sec2_problem}

\section{Method} \label{sec:method}
\input{sec3_method}

\input{sec4_experiments}

\section{Conclusions}
\input{conclusion}



{\small 
\bibliography{sigproc}
}

\end{document}

%% file: sec1_introduction.tex

Machine learning techniques combined with affordable sensors and processing power gave rise to {\em predictive maintenance} as a means to break the traditional trade-off between machine lifetime maximization (through corrective event-based maintenance) and downtime minimization (through preventive risk-based maintenance) \cite{maintenanceoverview}. 

Traditionally, predictive maintenance aims to schedule interventions on a machine based on health condition predictions derived from high frequency data collected by sensors. However, in the pharmaceutical industry, production machines must be audited by a medicines agency after each repair or maintenance operation. As a result, it is customary that maintenance and its associated auditing process are scheduled regularly (e.g., twice a year). In this context, predictive maintenance is no longer a regression problem: {\em when to schedule maintenance?} It becomes a classification problem: {\em what parts of a machine should be exchanged at a given point in time?}

We focus on one type of machine part: rolling-element bearings. Bearing products are important components of {\em active pharmaceutical ingredient} machinery, packaging machinery and medical testing equipment. The problem is how to predict the state of a bearing using vibration data. An existing approach is to consider a one class classification \cite{QIU2003127, HONG2014117} where a model is trained using only healthy bearing vibration data and the faults are detected when the signal deviates significantly from the healthy signal. Another approach is to train a supervised two class classifier using data from only the beginning of an experiment when a bearing is known to be healthy and the end of the same experiment when the bearing is known to be faulty \cite{s20205846}. However, binary classification does not provide the necessary level of bearing health condition granularity. The data can be segmented into multiple stages manually by inspecting plots of vibration signal throughout the whole bearing lifetime in both frequency and time domains \cite{manual, Kimotho2013MachineryPM}, or using clustering algorithms \cite{6798765, 7021915}.


In this paper, we build on the aforementioned ideas and present a framework for early stage detection of degradation in bearings. This is crucial to decide which bearings to replace at a given point in time. Specifically, our contributions are the following:
\begin{enumerate}
    \item We build on domain knowledge to automate data labeling. We propose a k-means bearing lifetime segmentation method based on high-frequency bearing vibration signal embedded in a latent low-dimensional subspace using an AutoEncoder.
    \item We design a supervised classifier, based on a multi-input neural network, for bearing degradation stage detection. 
    \item Our experiments with a publicly available dataset (see Section \ref{sec:experiments}) show that our framework is practical and efficient: (i) the  automatic labeling method is comparable to manual labeling, (ii) the classifier is trained quickly with high accuracy and (iii) it identifies stages reliably based on individual vertical and horizontal measurements.
\end{enumerate}



Code and experiments are available at \url{https://github.com/DovileDo/BearingDegradationStageDetection}.

%% file: sec5_related.tex
Existing work in the literature focuses on process validation in the pharmaceutical industry based on mathematical models~\cite{BANO2019254} rather than predictive maintenance based on sensor data. In the rest of this section, we focus on bearing degradation modeling.

Bearing degradation modeling methods can be categorized into two groups: continuous degradation, focused on building a single model to capture the degradation, and discrete degradation stage models, based on the assumption that the distribution governing the degradation changes over time. 

\textbf{Continuous degradation models} by nature are regression models that often rely on fault, commonly referred to as first predicting time (FPT), detection \cite{fpt}. Due to low healthy bearing vibration signal variation and high variation in bearing lifetime lengths, linear target function is set only after the FTP is detected and bearing vibration signal starts showcasing an exponential degradation trend. A health index for FPT detection is typically calculated from a single feature, such as RMS, or a combination of features derived from bearing vibration data and fused together, using e.g., PCA \cite{pca, pca2}, Mahalanobis distance \cite{wang2016two} or Chebyshev inequality function \cite{6783688}, is used as a health index for FPT detection. A fault is detected when the health index exceeds a predefined threshold value. This is a critical decision in the system, yet threshold value setting is not a trivial task and, at times, can be quite arbitrary. What is more, these methods are limited to a single health index that does not convey enough information to detect faults in early degradation stages. As a result, Continuous degradation models suffer from late fault detection.    

Unsupervised AutoEncoder based anomaly detection methods \cite{8651897,fan2017autoencoder} are used as an alternative to health index. Deep neural networks are able to extract rich representations of a healthy bearing that together with the decoded signal residuals are used for fault detection. In the proposed method, we leverage AutoEncoder based anomaly detection in the late bearing degradation stage. However, in the early degradation stages these methods face a similar challenge of an appropriate threshold setting that would allow early fault detection without increasing rate of false positives. Furthermore, as these methods are based on residuals, strictly only healthy bearing signals can be used for training which is hard to select and distinguish from early degradation stages. For these reasons, we do not use AutoEncoder based anomaly detection in the early stages of bearing degradation.

Wang et al. \cite{hybrid} estimated remaining useful life (RUL) directly by using exponential degradation models to fit the degradation curve composed of features extracted from a raw signal past measurements. Fault threshold setting is bypassed by extrapolating the degradation curve into the future and measuring the time until the extrapolated curve exceeds the complete failure threshold. This way, both health index building and threshold setting are avoided. Yet, due to a constant trend of the bearing degradation curve at early degradation stages, the extrapolation method fails to accurately estimate RUL early in a bearing's lifetime. 

Continuous degradation models by design are focused on modeling later stages of bearing degradation. Our work is different from these models as we attempt to build a framework for bearing condition monitoring throughout a bearing's lifetime. We therefore employ the Discrete degradation stage modeling method.

\textbf{Discrete degradation stage modeling} is a more robust time-varying bearing degradation process modeling approach compared to the Continuous degradation models, however faced with even harder challenge of identifying multiple degradation stages. 

Statistical models such as hidden Markov models \cite{6616438} or Bayesian belief networks \cite{6798765} are used to characterize (hidden) bearing degradation stages based on observations (vibration signals). These approaches consider low dimensional feature vector inputs since model parameter estimation in high dimensional settings is challenging \cite{10.2307/23566458}. Low dimensional inputs reduce model predictive power, as information is lost when bearing vibration signal is embedded in lower dimensions.

Liu et al. \cite{0954406215590167} model discrete degradation stages as a multi-class classification problem. Yet, they use low dimensional input vectors as well as an arbitrary number of classes that lead to low accuracy for early bearing degradation, both in terms of data labeling and classification. In contrast, our approach introduces and combines an automatic data labeling strategy based on domain knowledge with a deep neural network classifier that is well suited for predictions in both early and late degradation stages.

%% file: sec2_problem.tex

As a rolling-element bearing degrades, it goes through physical changes that can be detected in the frequency spectrum, and which can generally be divided into five categories:  

\begin{figure*}[h]
    \centering
    \includegraphics[width=1\textwidth]{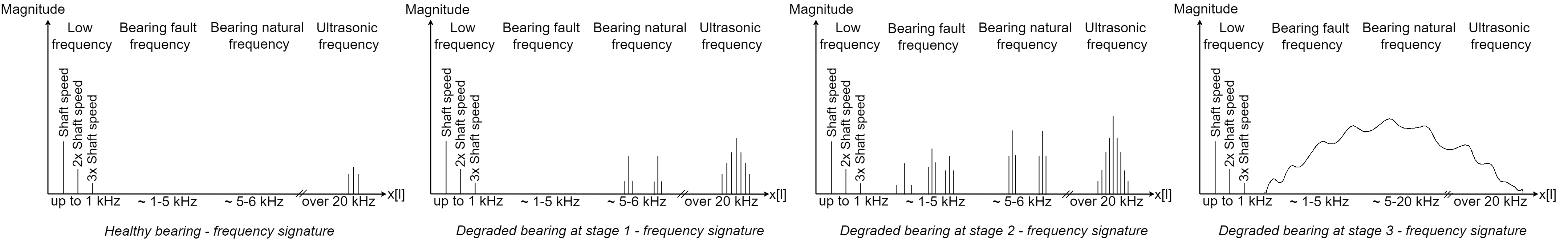}
    \caption[Bearing degradation stage signatures]{Bearing degradation stage signatures\footnotemark} \label{fig:stages}
\end{figure*}

\textbf{Healthy bearing.} While a bearing is healthy, fundamental frequency (shaft rotating speed) is the only frequency that appears in the frequency spectrum. Fundamental frequency persists throughout the bearing's lifetime. 

\textbf{Stage 0 - thinning lubrication.} The earliest sign of bearing deterioration are vibrations in the ultrasonic frequency region, indicating that bearing lubrication is beginning to thin, see Figure \ref{fig:stages}. In order to detect these signs, vibration data has to be sampled at a very high rate which may not be feasible in industrial applications. What is more, physical examination of a bearing at this stage may not reveal any signs of degradation. Therefore, throughout this paper, a bearing at this stage is considered to be healthy. 


\textbf{Stage 1 - insufficient lubrication.} Decreased lubrication left unmitigated leads to increased friction, causing the bearing to enter the second stage of degradation. Ultrasonic frequency vibrations continue to rise. In addition, changes appear in the natural frequency region as increased intensity of the impact forces agitates the natural frequencies of different bearing components, see Figure \ref{fig:stages}. At this stage, if disassembled, the bearing would have an insufficient level of lubrication.


\textbf{Stage 2 - bearing fault.} Rolling-element bearings are prone to four types of faults: outer race, inner race, rolling element, and cage fault~\cite{Zhu2014SurveyOC}. Due to a developed fault or a combination of faults, acceleration readings start to rise, and characteristic fault frequencies can be detected in a frequency range that we denote {\em bearing fault frequency}, see Figure \ref{fig:stages}. At this stage, the bearing should be immediately replaced. A specific bearing fault would be observed if the bearing was disassembled.


\textbf{Stage 3 - bearing failure.} When the bearing reaches the last deterioration stage, severe cracks appear and the edges of the raceway or rolling element defects get smoothed. This creates significant looseness in the bearing; therefore, high-frequency fault detection methods may decline as the characteristic fault frequencies are replaced by vibrations in the form of random noise, particularly in the lower frequency regions, see Figure \ref{fig:stages}. In this stage, the overall vibration velocity amplitudes rise significantly, and complete failure of the machinery can occur at any time. Ideally, no machinery should ever be pushed to this point.


\textbf{Problem.} Our key insight in this paper, is that a multi-class classification enables predictions at any point in time, based on a bearing's current condition.

Formally, the problem is the following: 
Given a discrete-time vibration signal $x[n]$  sampled at a high-frequency $n$ and obtained for a bearing at an unknown degradation stage $k$, where $k \in$ \{\textit{healthy bearing, stage 1, stage 2, stage 3}\}, the problem is to define a mapping function \begin{math}f: x[n] \mapsto k\end{math}.

We propose a method to automatically train a classifier that learns this mapping function.

\footnotetext{\url{https://www.stiweb.com/v/vspfiles/downloadables/appnotes/reb.pdf}}


%% file: sec3_method.tex

\begin{figure*}
    \centering
    \includegraphics[width=0.9\textwidth]{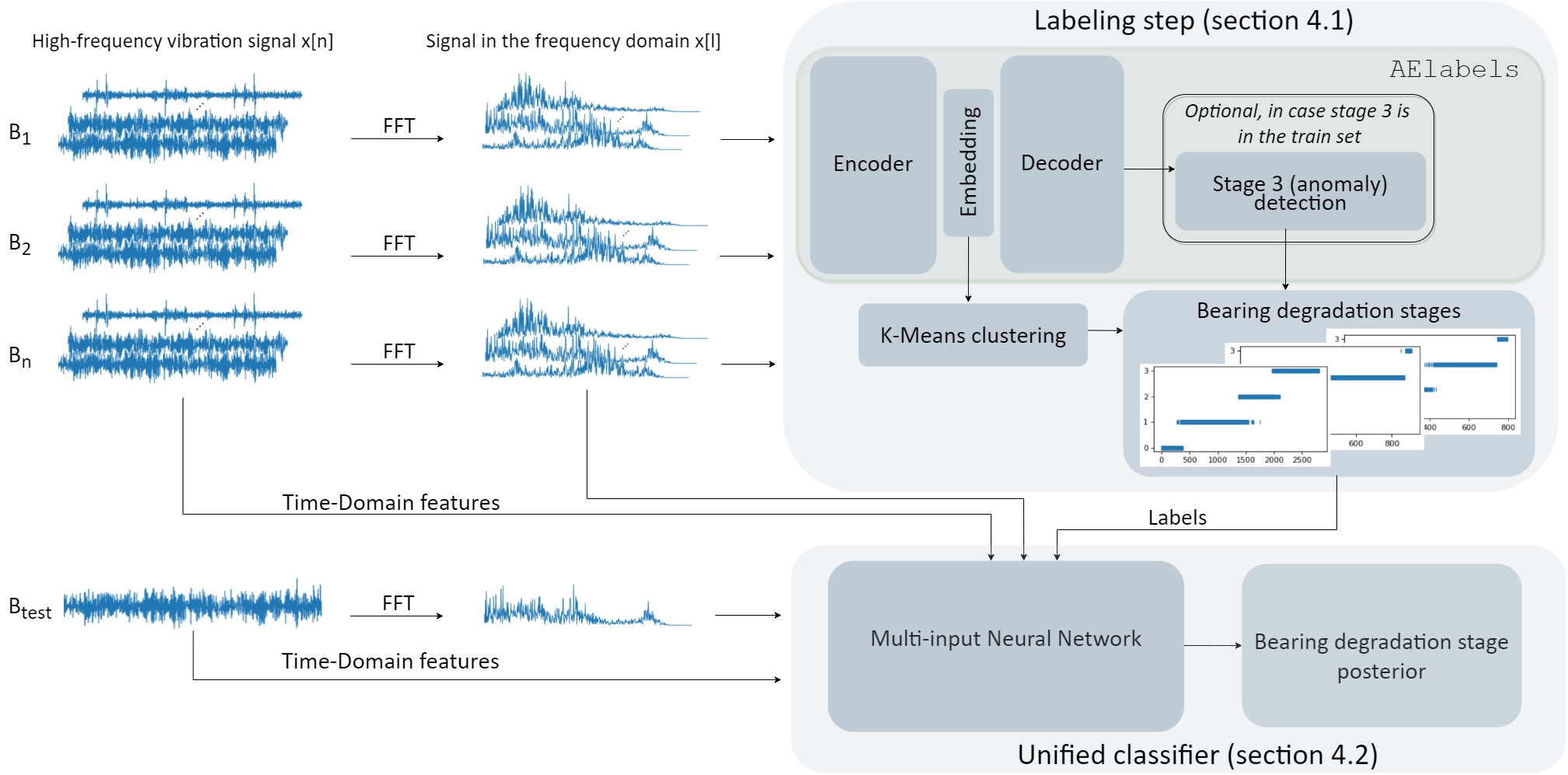}
    \caption{AutoEncoder based data labeling and bearing degradation stage detection \label{fig:1}}
\end{figure*}

We propose a two step method (see Figure \ref{fig:1} for an overview). In the first step, an unsupervised k-means clustering method is used to split a bearing's lifetime dataset into $k$ stages. Our assumption here is that experiments are performed on a range of representative bearings. The resulting datasets are automatically labeled within the pharmaceutical company, without the need for manual intervention from bearing experts.

The labeled data is used as input for the second step, where a unified classifier is trained to predict the stage of a vibration signal for a single bearing.

\subsection{Labeling} \label{labeling}

We consider bearing lifetime data, obtained by running experiments on healthy bearings until they fail. A bearing's lifetime dataset $D$ contains vibration signals for multiple bearings \{$B_1$, $B_2$, ..., $B_N$\}, where $B_i$ is a list of sequential bearing vibration signal vectors ($x[n]_{t0},$ $x[n]_{t1},$ ..., $x[n]_T$), obtained with sampling frequency $n$ for a fixed time at regular intervals, from the initial time $t_0$ at  which the bearing was healthy until the time $T$ where the bearing fails.

A common way to deal with unlabeled data is unsupervised clustering. In our case, unsupervised methods cannot be applied on multiple bearings at once. When trained on multiple bearings, unsupervised methods pick up differences in bearing fault modes (e.g., inner race outer race, rolling element, and cage fault), as well as different running speeds, or loads in addition to the degradation stages. To single out bearing degradation stages, every bearing in the training set is processed separately. 

The discrete-time vibration signal $x[n]$ is first transformed to the frequency domain using Fast Fourier Transform (FFT). As we are dealing with high-frequencies, the signal transformed to the frequency domain is high-dimensional, $x[n] \xrightarrow[]{\text{FFT}} x[l]$, where $n$ is sampling frequency and $l = n/2 + 1$. We consider two approaches for dimensionality reduction: AutoEncoder and principal component analysis (PCA).

\begin{figure}
    \centering
    \includegraphics[width=0.4\textwidth]{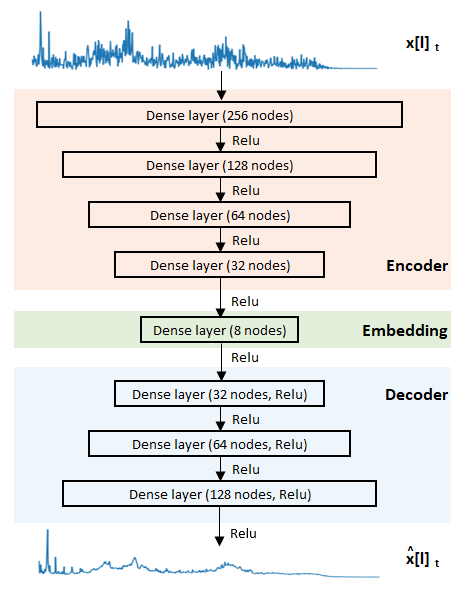}
    \caption{AutoEncoder architecture \label{fig:2}}
\end{figure}

First, we consider an AutoEncoder (architecture in Figure \ref{fig:2}) and apply k-means clustering to the data embedded in the latent low-dimensional space.  A separate AutoEncoder is trained for each bearing in the training set to reduce the mean absolute error (MAE) reconstruction loss:

\begin{equation*}
    \mathcal{L}_{MAE}(\theta) = \E \left\lVert X - \hat{X}(X, \theta)\right\rVert_1 = \frac{1}{T}\sum_{t=0}^{T} \left| X_t - \hat{X_t} \right| 
\end{equation*}

where $\theta$ is the AutoEncoder weights, \begin{math}X = \{x[l]\}_{t=0}^{T} \end{math}
are bearing vibration signals in the frequency domain from $t=0$ to $T$, and  \begin{math}\hat{X} = \{\hat{x}[l]\}_{t=0}^{T} \end{math} are AutoEncoder reconstructed bearing vibration signals.

The AutoEncoder serves two purposes. In addition to its ability to learn non-linear, complex data transformations for dimensionality reduction, it is also used as an anomaly detection method \cite{app8091468} for the degradation stage 3 labeling. Empirically, we observed that when stage 3 is present in the data, k-means clustering tend to split it into multiple stages and cluster early stages together as the variance within the stage 3 is much higher than the variance between the early degradation stages. Therefore, if it is known that stage 3 is in the data, we leave the end of a bearing lifetime (last 20\% of observations) out when training an AutoEncoder. Bearing degradation stage 3 is detected by analysing AutoEncoder decoded signal residuals. When residuals of a left out observation deviate by more than three standard deviations from the mean residual value of the reconstructed signals in the training dataset, then this observation is labeled as stage 3. The rest of the bearing's lifetime is labeled by clustering the data in the AutoEncoder embedding space using k-means.

PCA is often used as dimensionality reduction \cite{DONG20133143, PCA4304127}, as well as a basis for clustering \cite{0954406215590167, Wang_2015}. Therefore, we replaced the \texttt{AElabels} part of the method in Figure \ref{fig:1} to map vibration signals into lower dimensions. We observe that the first principal components explain only a small portion of the variance. We thus decided to keep the first 40 principal components of each signal (between 30\% and 60\% variance).

We compare the AutoEncoder and PCA methods for dimensionality reduction in Section \ref{sec:experiments}.


\subsection{Classifier}

\begin{figure}
    \centering
    \includegraphics[width=0.48\textwidth]{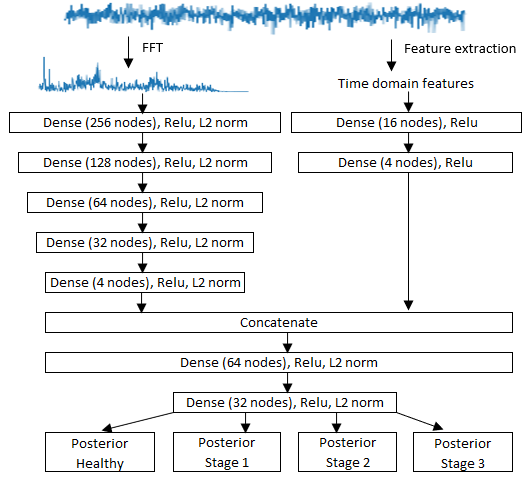}
    \caption{Classifier architecture \label{fig:3}}
\end{figure}

Earliest signs of bearing degradation are only visible in the frequency domain as described in Section \ref{sec:problem}. However, as degradation progresses, the signal in the time domain gets more and more informative. To capture both effects, our model architecture combines inputs in the time and frequency domains. Figure~\ref{fig:3} shows our proposed multi-input neural network, with bearing vibration signal input in both frequency and time domains.

Bearing vibration signal in the time domain is transformed to the frequency domain using FFT. A signal in the frequency domain can be further analyzed to obtain higher quality features using signal processing techniques such as spectral analysis, power spectrum analysis, power spectral density analysis and envelope analysis \cite{envelope}. Yet, this requires extensive mechanical engineering expertise. Neural networks are notorious for their ability to extract features automatically from relatively raw data. Therefore, the frequency domain input side of the neural network is used as a feature extractor first and then the extracted features are concatenated with time domain features. 

Most of the time domain features are motivated by the observation that samples of a healthy bearing vibration tend to follow a normal distribution, whereas distribution of the samples towards a bearing failure tend to deviate significantly from it. The thirteen time domain features are listed in Table \ref{table:features}. As Kullback–Leibler divergence (KL) is not symmetric, it makes up two features:  KL from the empirical distribution of an observation samples to \( \mathcal{N}(\Bar{x},\,s^{2})\) and KL from \(\mathcal{N}(\Bar{x},\,s^{2})\) to the empirical distribution. Skewness measures sample distribution asymmetry, kurtosis - distribution tail heaviness (outliers), and crest factor shows how extreme the signal peak is. Signal energy, RMS, number of peaks, and zero crossings tend to increase significantly towards a bearing failure point and are clear signs of total degradation. These features are commonplace in vibration signal analysis \cite{Zhu2014SurveyOC}.

\begin{table}
\centering
\caption{Time domain features}
\begin{tabular}{llll}
\hline
1.  & \begin{tabular}[c]{@{}l@{}}Mean\\ \( \frac{1}{n}\sum_i^n x_i\)\end{tabular} & 
2.  & \begin{tabular}[c]{@{}l@{}}Abs. median  \\ \( \text{Median}(|x_i|,...,|x_n|)\)\end{tabular}               \\
\\
3.  & \begin{tabular}[c]{@{}l@{}}Standard deviation\\ \( \sqrt{\frac{1}{n} \sum_i^n(x_i - \Bar{x}_n})^2 \)\end{tabular} &
4.  & \begin{tabular}[c]{@{}l@{}}Skewness \\ \( \frac{\frac{1}{n} \sum_i^n (x_i - \Bar{x}_n)^3}{(\frac{1}{n} \sum_i^n (x_i - \Bar{x}_n)^2)^\frac{3}{2}} \)\end{tabular}                   \\
\\
5.  & \begin{tabular}[c]{@{}l@{}}Kurtosis \\ \( \frac{\frac{1}{n} \sum_i^n (x_i - \Bar{x}_n)^4}{(\frac{1}{n} \sum_i^n (x_i - \Bar{x}_n)^2)^2} \)\end{tabular}       & 6.  & \begin{tabular}[c]{@{}l@{}}Crest factor \\ \( \frac{\text{max}|x|} {\sqrt{\frac{1}{n} \sum_i^n x_i^2}}\)\end{tabular}                      \\
\\
7.  & \begin{tabular}[c]{@{}l@{}}Energy \\ \(\sum_i^n |x_i|^2 \)\end{tabular}     & 
8.  & \begin{tabular}[c]{@{}l@{}}RMS  \\ \( \sqrt{\frac{1}{n} \sum_i^n x_i^2} \)\end{tabular}       \\
\\
9.  & \# of peaks & 10. & \# of zero crossings                      \\
\\
11. & Shapiro test                                                       & 12. & KL divergence
\\
\hline
\end{tabular}
\label{table:features}
\end{table}

The neural network is trained to optimise categorical cross-entropy loss:

\begin{equation*}
    \mathcal{L}_{CE}(\theta) = - \sum_{c=1}^C k_c \log(y_c)
\end{equation*}

where $C \leq 4$ is the number of bearing degradation stages, $k$ is the bearing degradation stage vector predicted by the k-means clustering as described in Subsection \ref{labeling}, and $y = f(x[l],x[n]')$ is the bearing degradation stage posterior probability vector predicted by the classifier.

It is worth noting that our multi-input architecture design leverages reduced input dimensionality and thus offers computational advantages, in addition to its ability to analyse bearing vibration signal in both frequency and time domains.  

%% file: sec4_experiments.tex
\section{Experimental Framework} \label{sec:experiments}

Our goal is to experimentally explore the viability of our framework: how well can degradation stages be distinguished automatically for the training set? How well can the stage a bearing is in be predicted using a vibration signal? In this section, we describe the experimental framework. We present our results in the next section. All experiments were executed on a Lenovo ThinkPad X390 touch 13 Core i 5 16GB RAM, 512 GB SSD.

\subsection{System}
The proposed framework was implemented in Python. Pandas \cite{mckinney-proc-scipy-2010} and NumPy \cite{harris2020array} were used for data preprocessing and extraction of some time domain features. The remaining time domain features and all frequency domain features were extracted using signal processing functions in the SciPy \cite{2020SciPy-NMeth} library. The Scikit-learn \cite{scikit-learn} implementation of k-means was used for clustering and PCA dimensionality reduction. Both AutoEncoders and the multi-input neural network were implemented with Keras \cite{chollet2015keras}. The labeling portion of the framework is 331 SLOC, and the classifier is 51 SLOC.

\subsection{Workload}
The proposed framework is trained and evaluated using the FEMTO Bearing dataset\footnote{\url{https://ti.arc.nasa.gov/tech/dash/groups/pcoe/prognostic-data-repository/#femto}}, which was presented in the IEEE PHM 2012 Prognostic challenge~\cite{nasa2}. 

This dataset contains horizontal and vertical bearing vibration measurements collected by running experiments on bearings. To accelerate the degradation process, bearings were put under stress conditions exceeding the recommended load force. A total of 17 bearings were tested under three different load and rotational speed conditions, detailed in table \ref{table:conditions}. Due to safety concerns, experiments were stopped after the accelerometer readings exceeded 20 g, signifying that the bearing reached the final degradation stage. 

\begin{figure}
    \centering
    \includegraphics[width=0.38\textwidth]{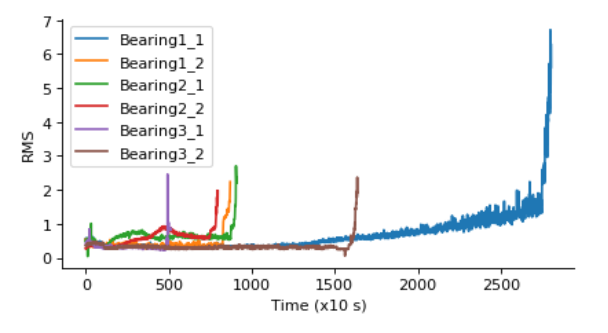}
    \caption{Lifetimes of training set bearings. The x-axis shows the time ($\times$ 10 seconds), the y-axis shows the root mean square (RMS) of the horizontal vibration acceleration. Sharp increase in RMS indicates failure, however early stages of degradation can not be predicted from the RMS alone.}
    \label{fig:rms}
\end{figure}

The FEMTO Bearing dataset is challenging due to high variability in bearing lifetime durations (from 28 minutes to 7 hours), see Figure \ref{fig:rms} for training set bearings. Bearing faults are not specified. Worse, it is noted in the data challenge description that bearings may have suffered from multiple defects at once. 

During the experiments, bearing vibrations were sampled every 10 s at 25,600 Hz for 0.1 s. Sampling at such a high frequency in industrial applications may not be feasible. So, we downsampled the raw vibration signal of 25,600 Hz in the FEMTO dataset by half to 12,800 Hz. This significantly reduces data dimensionality yet preserves an actionable frequency range. After the transformation to the frequency domain, we obtain frequencies up to 6,400 Hz, where the signs of bearing degradation are expected to appear.


\begin{table}[t]
\centering
\caption{Bearing operating conditions and dataset split}
\begin{tabular}{c|cccc}
\hline
Condition & Speed     & Load    & \multicolumn{2}{c}{\# of bearings in} \\
\#        & rpm       & N       & train set          & test set         \\ \hline
1         & 1,800 rpm & 4,000 N & 2                  & 5                \\
2         & 1,650 rpm & 4,200 N & 2                  & 5                \\
3         & 1,500 rpm & 5,000 N & 2                  & 1                \\ \hline
\end{tabular}
\label{table:conditions}
\end{table}
Downsampled signal in the frequency domain is represented by 641 features (12,800 Hz (sampling rate) x 0.1 s (sampling duration) / 2 (due to the Sampling Theorem) + 1 (zero frequency)). The dataset contains both vertical and horizontal vibration signals; therefore, both signals were transformed to the frequency domain.

\subsection{Metrics}
\label{sec:metrics}
In the absence of ground truth, we need to introduce metrics to characterize (a) how well stages are distinguished in the labeling phase, and (b) how well the classifier predicts the bearing stage given a vibration signal.

\begin{figure}[t]
    \centering
    \includegraphics[width=0.5\textwidth]{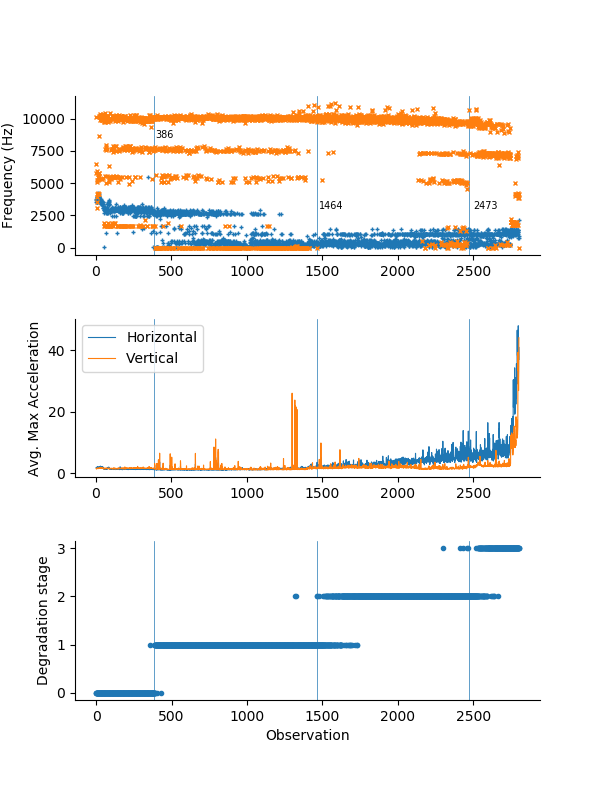}
    \caption{Bearing 1\_1 labeling. Top: highest magnitude frequency of each observation of both horizontal and vertical vibration signals. Middle: smoothed maximum acceleration calculated by averaging five highest absolute acceleration measurements in the time domain. The top and middle graphs are used for manual segmentation (vertical lines).  Bottom:  bearing 1\_1 with \texttt{AElabels}. Manual and automatic labels largely overlap (high accuracy).  \label{fig:5}}
\end{figure}

To evaluate labeling, we compare the automatically generated stage labels with the ones that we created manually using the method from~\cite{manual} (for the training set). The less discrepancy the better, so we measure accuracy.  Figure~\ref{fig:5} illustrates this metric.  

To evaluate the classifier, we first measure the time it takes to train the classifier. We also compare the predicted stage with the ones we labeled by the k-means clustering (for the test set). For predictive maintenance, it is imperative that predictions, at various points in time, follow the sequence of degradation stages. So, in addition to accuracy at random points in time, we also measure how well the predicted stages overlap with the actual stages when predictions are made sequentially in time.

\subsection{Experiments}

We split the FEMTO Bearing dataset into a training set and a test set, see Table~\ref{table:conditions}.
Observations of all six bearings in the training set were labeled automatically using both the AutoEncoder and PCA methods. As the FEMTO dataset contains measurements of horizontal and vertical vibrations, both signals were used as inputs for labeling. Separate AutoEncoders were trained for each bearing, for both vertical and horizontal vibration signals. In total, 12 AutoEncoders were trained. Horizontal and vertical vibration signals embedded by the AutoEncoders were concatenated, resulting in a latent space of 16 dimensions where k-means clustering was applied. As it is known that all bearings during the experiments reached stage 3, the last 20\% of observations of each bearing were left out when training the AutoEncoders. Stage 3 was labeled using horizontal vibration signal anomaly detection and the rest of the stages were labeled using 3-means clustering. We dub labels generated by the described method \texttt{AElabels}.

We replaced the \texttt{AElabels} part of the labeling method in Figure \ref{fig:1} with PCA to generate \texttt{PCAlabels} for reference. Again, both horizontal and vertical signals reduced to 40 dimensions were concatenated and clustered using 4-means clustering. 

The data automatically labeled with \texttt{AElabels} and \texttt{PCAlabels} was used to train two instances of the classifier that takes horizontal and vertical vibration signals as input vectors. 

Our experiments are based on the test set from the FEMTO Bearing dataset. They focus on (1) the accuracy of the automatic labeling method, (2) the time to train the classifier and (3) the accuracy of the predictions. Throughout, we compare the impact of the dimensionality reduction methods \texttt{AElabels} and \texttt{PCAlabels}.

\section{Experimental Results}

\subsection{Data labeling}

Figure \ref{fig:5} illustrates the accuracy of \texttt{AElabels} for a single bearing. The automatic method is largely aligned with manual labeling, with some overlap across neighbouring stages.

\begin{figure}
    \centering
    \includegraphics[width=0.48\textwidth]{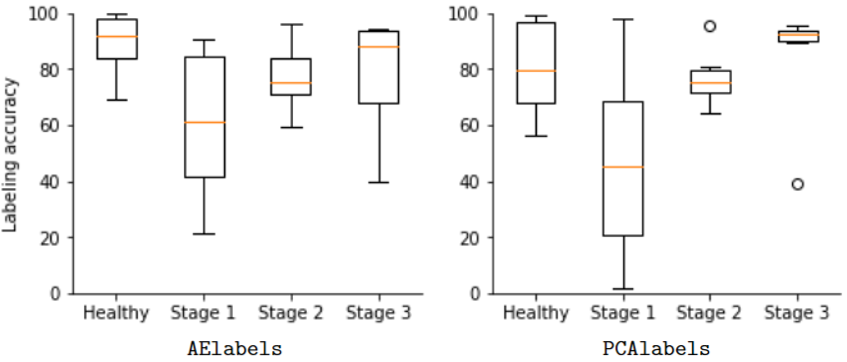}
    \caption{Bearing degradation stage labeling, as described in Subsection \ref{labeling}, accuracy. Left: \texttt{AElabels} vs manual labeling. Right: \texttt{PCAlabels} vs manual labeling. Both labeling methods lead to high accuracies overall, except for stage 1.  \label{fig:4}}
\end{figure}

Figure \ref{fig:4} is a box plot that shows the distribution of accuracy across all bearings in the train set. It shows the accuracy of the \texttt{AElabels} and \texttt{PCAlabels} methods, for each stage. Both labeling methods perform well for all bearings except bearing 1\_2. It is worth noting that manual labeling of bearing 1\_2 was challenging as frequency and acceleration graphs did not share the general degradation patterns of other train set bearings. Also, the separation between healthy bearing and degradation stage 1 can at times be arbitrary when labeling manually. Therefore, low accuracy is likely caused by incorrect manual labeling.

\subsection{Classifier}

\begin{figure}[h]
    \centering
    \includegraphics[width=0.35\textwidth]{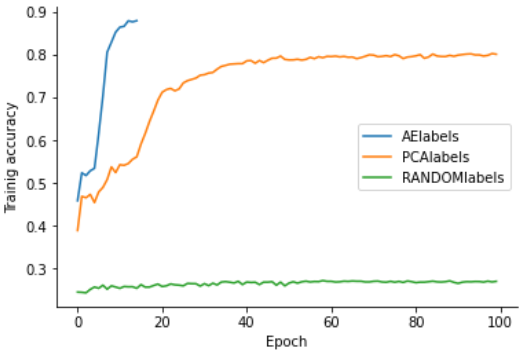}
    \caption{Classifier training accuracy.\label{fig:trainingacc}}
\end{figure}

\textbf{Training.} Figure~\ref{fig:trainingacc} shows the classifier training accuracy. Classifier training on \texttt{PCAlabels} takes more epochs and it reaches lower accuracy than the classifier trained on \texttt{AElabels}. This is a very significant advantage in terms of practicality for the \texttt{AElabels} method, as it has the potential to scale with the number of bearings and machines in production.

As a sanity check, a classifier trained on randomly generated labels reaches around 25\% training accuracy which is expected for a four class classification.


\begin{figure}[t]
    \centering
    \includegraphics[width=0.48\textwidth]{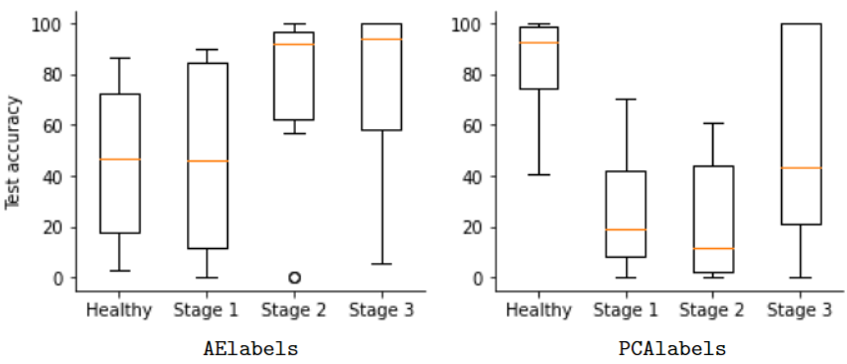}
    \caption{Classifier test set prediction accuracy. Left: classifier trained on \texttt{AElabels} vs test set \texttt{AElabels}. Right: classifier trained on \texttt{PCAlabels} vs test set \texttt{PCAlabels}.   \label{fig:testacc}}
\end{figure}

\begin{figure}[t]
    \centering
    \includegraphics[width=0.48\textwidth]{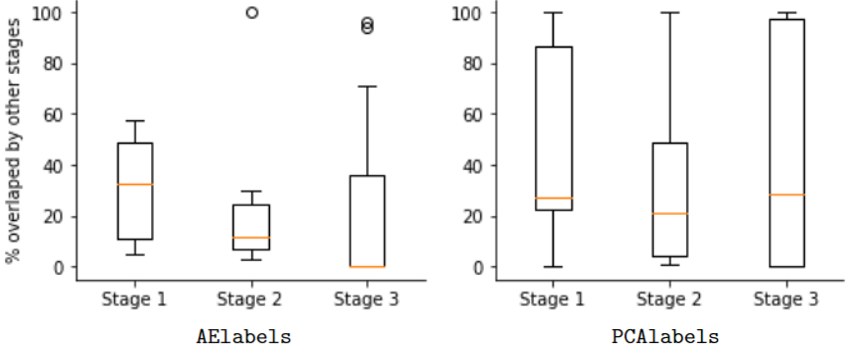}
    \caption{Percent of stage length (between first and last prediction of the given class) overlapped by any other stage.  \label{fig:overlap}}
\end{figure}

\begin{figure*}[h]
    \centering
    \includegraphics[width=0.98\textwidth]{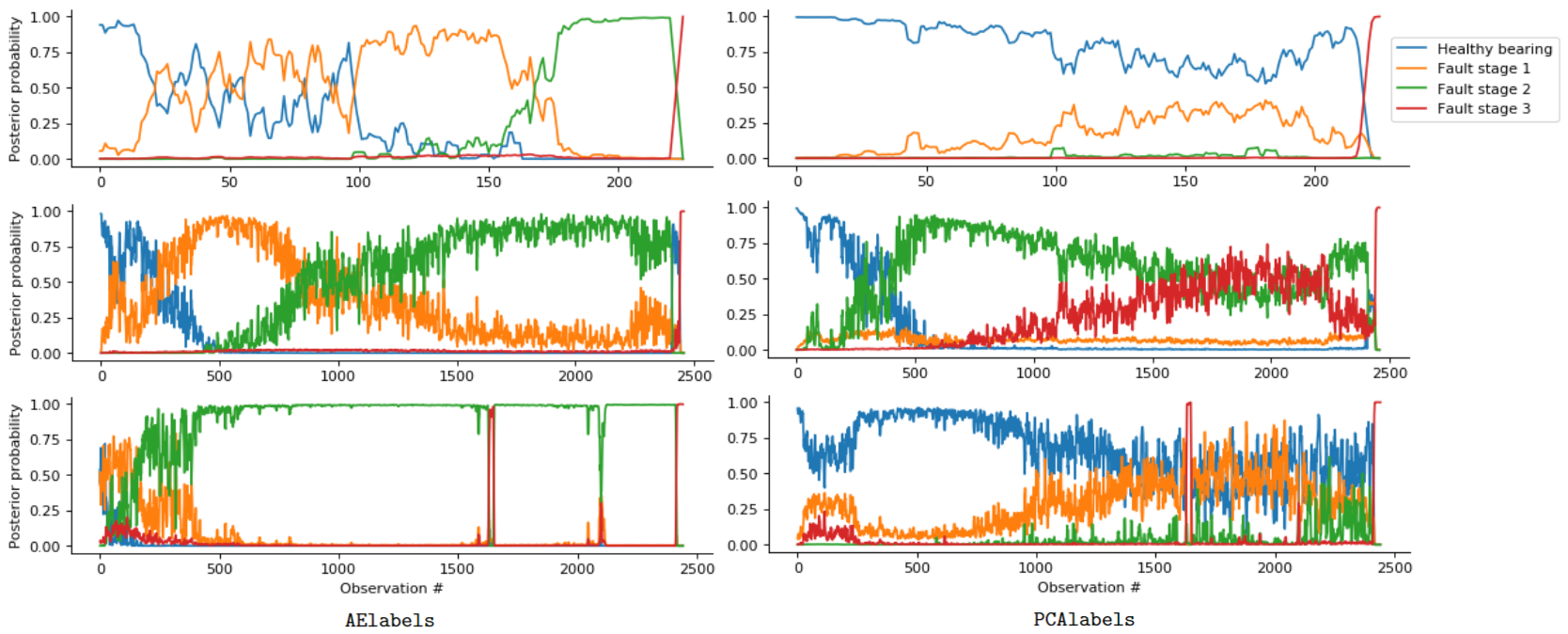}
    \caption{Classifier predictions for bearings 2\_7 (top), 1\_5 (middle) and 1\_6 (bottom). Left: predictions of the classifier trained on \texttt{AElabels}. Right: predictions of the classifier trained on \texttt{PCAlabels}.  \label{fig:predictions}}
\end{figure*}

\textbf{Inference.} As explained in Section~\ref{sec:metrics}, predictions were generated for all bearings in the test set sequentially in time. Classifier predictions were smoothed by averaging the posterior probability of the five most recent predictions.

Figure~\ref{fig:predictions} illustrates the posterior probabilities of bearing degradation stages predicted for selected bearings by the classifiers trained with \texttt{AElabels} and \texttt{PCAlabels}. The training set contains only two bearings with lifetimes longer than 4 hours, while four bearings' lifetimes are shorter than 3 hours. These differences in lifetime length are likely influenced by different fault modes. We selected typical predictions for bearings with both long and short lifetimes. 
The \texttt{AElabels} classifier generates predictions that neatly fall into one of the four stages. Basically, the four stages can be reconstructed from the predictions. Recall that each prediction is based on a single vibration signal without historical context.

The \texttt{PCAlabels} classifier is less accurate. For instance, the top graph shows stage 1 prediction throughout the lifetime of the bearing. This is a liability for predictive maintenance. Also, the middle graph shows instability in time, with stage 2 predicted early and late in the bearing's lifetime. The bottom graph shows a fault that is detected too early with \texttt{AElabels} and way too late with \texttt{PCAlabels}. The former prediction is conservative and thus well-suited for predictive maintenance in the context of this study.

Figures~\ref{fig:testacc} and \ref{fig:overlap} show the classifier prediction accuracy (with respect to k-means clustering labels) and the percentage of a stage length that overlaps with other stages across all bearings. The classifier trained on \texttt{AElabels} reaches much higher accuracy in later degradation stages. Bearing degradation stages predicted by the classifier trained on \texttt{AElabels} overlap less than those trained with \texttt{PCAlabels}. The classifier trained on \texttt{AElabels} predicts healthy bearing (healthy or stage 1) less often thus supporting a more conservative predictive maintenance approach.

Finally, we explore the quality of the decisions that can be made based on the classifier's predictions. We consider that a bearing fault is identified when the classifier predicted stage 2 or stage 3, whichever happens first, for the first time.
Table~\ref{table:decision} shows the timing of the faults detected by the classifiers trained on both \texttt{AElabels} and \texttt{PCAlabels} compared to (a) the remaining healthy lifetime after the fault is detected (the smaller the better) and (b) the remaining total bearing lifetime remaining after the fault is detected (it should be high enough to guarantee that non healthy parts are picked for replacement when maintenance is scheduled). Again, the \texttt{AElabels} method performs well regardless of the bearing lifetime. In contrast, the \texttt{PCAlabels} method tends to lead to late detection decisions when the lifetime of a bearing is short.

\begin{table}[h]
\centering
\caption{Bearing faults detected by the classifiers trained on \texttt{AElabels} and \texttt{PCAlabels}. The asterisk (*) indicates cases where faults were detected either too early ($>$90\% of the lifetime left), or too late ($<$10\% of the lifetime).}
\small
\begin{tabular}{c|rr|rr|r}
\hline
Bearing               & \multicolumn{2}{c|}{\% of healthy} & \multicolumn{2}{c|}{\% of lifetime}  & \multicolumn{1}{c}{Total} \\
ID                    & \multicolumn{2}{c|}{after fault}      & \multicolumn{2}{c|}{left after fault} & \multicolumn{1}{c}{length}   \\ \hline
\multicolumn{1}{l|}{} & \scriptsize{\texttt{AElabels}}               & \scriptsize{\texttt{PCAlabels}}               & \scriptsize{\texttt{AElabels}}            & \scriptsize{\texttt{PCAlabels}}           & \multicolumn{1}{l}{}         \\ \hline
1\_3                  & 4.2                   & 11.0                   & 47.4                & 61.5                & 2371                         \\
1\_4                  & 0.0                   & 20.1                   & 23.7                & 58.4                & 1424                         \\
1\_5                  & 8.1                   & 13.7                   & 69.0                & 90.2*               & 2459                         \\
1\_6                  & 4.6                   & 97.2                   & 93.2*               & 32.2                & 2444                         \\
1\_7                  & 42.4                  & 22.6                   & 68.2                & 99.5*               & 2255                         \\
2\_3                  & 15.6                  & 28.8                   & 87.0                & 86.9                & 1951                         \\
2\_4                  & 31.9                  & 0.0                    & 34.0                & 0.8*                & 747                          \\
2\_5                  & 20.8                  & 35.6                   & 99.8*               & 100.0*              & 2307                         \\
2\_6                  & 3.1                   & 8.9                    & 45.2                & 76.0                & 697                          \\
2\_7                  & 0.0                   & 0.0                    & 27.9                & 3.1*                & 226                          \\
3\_3                  & 11.5                  & 0.0                    & 80.2                & 95.6*               & 430                          \\ \hline
\end{tabular}
\label{table:decision}
\end{table}

%% file: conclusion.tex
We focused on the problem of predictive maintenance in the pharmaceutical industry, where the issue is not when to schedule maintenance, but which parts of a machine to replace at a given point in time. We proposed a framework for predicting the degradation stages of rolling-element bearings, which are a key component of {\em active pharmaceutical ingredient} machinery, packaging machinery, and medical testing equipment. This framework is based on (1) {\em automatic labeling}: a k-means bearing lifetime segmentation method based on high-frequency bearing vibration signal embedded in a latent low-dimensional subspace using an AutoEncoder, and (2) {\em a multi-class classifier}: a multi-input neural network taking a 2-dimensional vibration signal as input and predicting the degradation stage.
Our experiments with the FEMTO dataset gave evidence that our approach is promising as it scales well (thanks to the automated labeling and the short training time) and produces actionable predictions.

A lot of work remains to be done before this method can be deployed in production. First, our method assumes that a training set is obtained from representative bearings. Whether a zero-shot learning approach is possible in this domain is an open question, however. It would enable a straightforward application to a wide range of different bearings in production. Even if zero-shot learning is not possible, further work is needed to characterize the scope of application of a given degradation stage model. Finally, a deployment of our method requires that vibration signals can be obtained cheaply and reliably from a large number of bearings in production. Whether these measurements can be obtained from embedded sensors deployed permanently (and thus accredited by the regulator) or from sensors deployed manually before scheduled maintenance periods is another open question.

%% file: main.bbl
\begin{thebibliography}{10}

\bibitem{app8091468}
T.~Amarbayasgalan, B.~Jargalsaikhan, and K.~H. Ryu.
\newblock Unsupervised novelty detection using deep autoencoders with density
  based clustering.
\newblock {\em Applied Sciences}, 8(9), 2018.

\bibitem{BANO2019254}
G.~Bano, P.~Facco, M.~Ierapetritou, F.~Bezzo, and M.~Barolo.
\newblock Design space maintenance by online model adaptation in pharmaceutical
  manufacturing.
\newblock {\em Computers \& Chemical Engineering}, 127:254--271, 2019.

\bibitem{chollet2015keras}
F.~Chollet et~al.
\newblock Keras, 2015.

\bibitem{DONG20133143}
S.~Dong and T.~Luo.
\newblock Bearing degradation process prediction based on the pca and optimized
  ls-svm model.
\newblock {\em Measurement}, 46(9):3143--3152, 2013.

\bibitem{fan2017autoencoder}
J.~Fan, W.~Wang, and H.~Zhang.
\newblock Autoencoder based high-dimensional data fault detection system.
\newblock In {\em 2017 ieee 15th international conference on industrial
  informatics (indin)}, pages 1001--1006. IEEE, 2017.

\bibitem{pca2}
T.~Gao, Y.~Li, X.~Huang, and C.~Wang.
\newblock Data-driven method for predicting remaining useful life of bearing
  based on bayesian theory.
\newblock {\em Sensors}, 21:182, 12 2020.

\bibitem{harris2020array}
C.~R. Harris, K.~J. Millman, S.~J. van~der Walt, R.~Gommers, P.~Virtanen,
  D.~Cournapeau, E.~Wieser, J.~Taylor, S.~Berg, N.~J. Smith, R.~Kern, M.~Picus,
  S.~Hoyer, M.~H. van Kerkwijk, M.~Brett, A.~Haldane, J.~F. del R{\'{i}}o,
  M.~Wiebe, P.~Peterson, P.~G{\'{e}}rard-Marchant, K.~Sheppard, T.~Reddy,
  W.~Weckesser, H.~Abbasi, C.~Gohlke, and T.~E. Oliphant.
\newblock Array programming with {NumPy}.
\newblock {\em Nature}, 585(7825):357--362, Sept. 2020.

\bibitem{HONG2014117}
S.~Hong, Z.~Zhou, E.~Zio, and W.~Wang.
\newblock An adaptive method for health trend prediction of rotating bearings.
\newblock {\em Digital Signal Processing}, 35:117--123, 2014.

\bibitem{7021915}
K.~Javed, R.~Gouriveau, and N.~Zerhouni.
\newblock A new multivariate approach for prognostics based on extreme learning
  machine and fuzzy clustering.
\newblock {\em IEEE Transactions on Cybernetics}, 45(12):2626--2639, 2015.

\bibitem{Kimotho2013MachineryPM}
J.~K. Kimotho, C.~Sondermann-Woelke, T.~Meyer, and W.~Sextro.
\newblock Machinery prognostic method based on multi-class support vector
  machines and hybrid differential evolution – particle swarm optimization.
\newblock {\em Chemical engineering transactions}, 33:619--624, 2013.

\bibitem{fpt}
Y.~Lei, N.~Li, L.~Guo, N.~Li, T.~Yan, and J.~Lin.
\newblock Machinery health prognostics: A systematic review from data
  acquisition to rul prediction.
\newblock {\em Mechanical Systems and Signal Processing}, 104:799--834, 2018.

\bibitem{0954406215590167}
Z.~Liu, M.~J. Zuo, and Y.~Qin.
\newblock Remaining useful life prediction of rolling element bearings based on
  health state assessment.
\newblock {\em Proceedings of the Institution of Mechanical Engineers, Part C:
  Journal of Mechanical Engineering Science}, 230(2):314--330, 2016.

\bibitem{envelope}
M.~S.~R. Mohd~Saufi, Z.~Ahmad, M.~Lim, and M.~Leong.
\newblock A review on signal processing techniques for bearing diagnostics.
\newblock {\em International Journal of Mechanical Engineering and Technology},
  8:327--337, 01 2017.

\bibitem{nasa2}
P.~Nectoux, R.~Gouriveau, K.~Medjaher, E.~Ramasso, B.~Morello, N.~Zerhouni, and
  C.~Varnier.
\newblock Pronostia: An experimental platform for bearings accelerated life
  test.
\newblock {\em IEEE International Conference on Prognostics and Health
  Management}, 2012.

\bibitem{scikit-learn}
F.~Pedregosa, G.~Varoquaux, A.~Gramfort, V.~Michel, B.~Thirion, O.~Grisel,
  M.~Blondel, P.~Prettenhofer, R.~Weiss, V.~Dubourg, J.~Vanderplas, A.~Passos,
  D.~Cournapeau, M.~Brucher, M.~Perrot, and E.~Duchesnay.
\newblock Scikit-learn: Machine learning in {P}ython.
\newblock {\em Journal of Machine Learning Research}, 12:2825--2830, 2011.

\bibitem{8651897}
E.~Principi, D.~Rossetti, S.~Squartini, and F.~Piazza.
\newblock Unsupervised electric motor fault detection by using deep
  autoencoders.
\newblock {\em IEEE/CAA Journal of Automatica Sinica}, 6(2):441--451, 2019.

\bibitem{6783688}
Y.~Qian, R.~Yan, and S.~Hu.
\newblock Bearing degradation evaluation using recurrence quantification
  analysis and kalman filter.
\newblock {\em IEEE Transactions on Instrumentation and Measurement},
  63(11):2599--2610, 2014.

\bibitem{QIU2003127}
H.~Qiu, J.~Lee, J.~Lin, and G.~Yu.
\newblock Robust performance degradation assessment methods for enhanced
  rolling element bearing prognostics.
\newblock {\em Advanced Engineering Informatics}, 17(3):127--140, 2003.
\newblock Intelligent Maintenance Systems.

\bibitem{maintenanceoverview}
Y.~Ran, X.~Zhou, P.~Lin, Y.~Wen, and R.~Deng.
\newblock A survey of predictive maintenance: Systems, purposes and approaches,
  2019.

\bibitem{PCA4304127}
L.~Shuang and L.~Meng.
\newblock Bearing fault diagnosis based on pca and svm.
\newblock In {\em 2007 International Conference on Mechatronics and
  Automation}, pages 3503--3507, 2007.

\bibitem{6616438}
F.~Sloukia, M.~E. Aroussi, H.~Medromi, and M.~Wahbi.
\newblock Bearings prognostic using mixture of gaussians hidden markov model
  and support vector machine.
\newblock In {\em 2013 ACS International Conference on Computer Systems and
  Applications (AICCSA)}, pages 1--4, 2013.

\bibitem{10.2307/23566458}
N.~Städler and S.~Mukherjee.
\newblock Penalized estimation in high-dimensional hidden markov models with
  state-specific graphical models.
\newblock {\em The Annals of Applied Statistics}, 7(4):2157--2179, 2013.

\bibitem{s20205846}
S.~Suh, J.~Jang, S.~Won, M.~S. Jha, and Y.~O. Lee.
\newblock Supervised health stage prediction using convolutional neural
  networks for bearing wear.
\newblock {\em Sensors}, 20(20), 2020.

\bibitem{manual}
E.~Sutrisno, H.~Oh, A.~S.~S. Vasan, and M.~Pecht.
\newblock Estimation of remaining useful life of ball bearings using data
  driven methodologies.
\newblock In {\em 2012 IEEE Conference on Prognostics and Health Management},
  pages 1--7, 2012.

\bibitem{2020SciPy-NMeth}
P.~Virtanen, R.~Gommers, T.~E. Oliphant, M.~Haberland, T.~Reddy, D.~Cournapeau,
  E.~Burovski, P.~Peterson, W.~Weckesser, J.~Bright, S.~J. {van der Walt},
  M.~Brett, J.~Wilson, K.~J. Millman, N.~Mayorov, A.~R.~J. Nelson, E.~Jones,
  R.~Kern, E.~Larson, C.~J. Carey, {\.I}.~Polat, Y.~Feng, E.~W. Moore,
  J.~{VanderPlas}, D.~Laxalde, J.~Perktold, R.~Cimrman, I.~Henriksen, E.~A.
  Quintero, C.~R. Harris, A.~M. Archibald, A.~H. Ribeiro, F.~Pedregosa, P.~{van
  Mulbregt}, and {SciPy 1.0 Contributors}.
\newblock {{SciPy} 1.0: Fundamental Algorithms for Scientific Computing in
  Python}.
\newblock {\em Nature Methods}, 17:261--272, 2020.

\bibitem{hybrid}
B.~Wang, Y.~Lei, N.~Li, and N.~Li.
\newblock A hybrid prognostics approach for estimating remaining useful life of
  rolling element bearings.
\newblock {\em IEEE Transactions on Reliability}, 69(1):401--412, 2020.

\bibitem{Wang_2015}
F.~Wang, J.~Sun, D.~Yan, S.~Zhang, L.~Cui, and Y.~Xu.
\newblock A feature extraction method for fault classification of rolling
  bearing based on {PCA}.
\newblock {\em Journal of Physics: Conference Series}, 628:012079, jul 2015.

\bibitem{pca}
T.~Wang.
\newblock Bearing life prediction based on vibration signals: A case study and
  lessons learned.
\newblock In {\em 2012 IEEE Conference on Prognostics and Health Management},
  pages 1--7, 2012.

\bibitem{wang2016two}
Y.~Wang, Y.~Peng, Y.~Zi, X.~Jin, and K.-L. Tsui.
\newblock A two-stage data-driven-based prognostic approach for bearing
  degradation problem.
\newblock {\em IEEE Transactions on industrial informatics}, 12(3):924--932,
  2016.

\bibitem{mckinney-proc-scipy-2010}
{W}es {M}c{K}inney.
\newblock {D}ata {S}tructures for {S}tatistical {C}omputing in {P}ython.
\newblock In {S}t\'efan van~der {W}alt and {J}arrod {M}illman, editors, {\em
  {P}roceedings of the 9th {P}ython in {S}cience {C}onference}, pages 56 -- 61,
  2010.

\bibitem{6798765}
X.~Zhang, J.~Kang, and T.~Jin.
\newblock Degradation modeling and maintenance decisions based on bayesian
  belief networks.
\newblock {\em IEEE Transactions on Reliability}, 63(2):620--633, 2014.

\bibitem{Zhu2014SurveyOC}
J.~Zhu, T.~Nostrand, C.~Spiegel, and B.~Morton.
\newblock Survey of condition indicators for condition monitoring systems.
\newblock {\em PHM 2014 - Proceedings of the Annual Conference of the
  Prognostics and Health Management Society 2014}, pages 635--647, 01 2014.

\end{thebibliography}
